\pdfoutput=1

\documentclass[11pt]{article}

\usepackage[final]{acl}

\usepackage{times}
\usepackage{latexsym}

\usepackage[T1]{fontenc}

\usepackage[utf8]{inputenc}

\usepackage{microtype}

\usepackage{inconsolata}

\usepackage{graphicx}

\title{TokenShapley: Token Level Context Attribution with Shapley Value}

\author{
 \textbf{Yingtai Xiao\textsuperscript{1}},
 \textbf{Yuqing Zhu\textsuperscript{1}},
 \textbf{Sirat Samyoun\textsuperscript{1}},
 \textbf{Wanrong Zhang\textsuperscript{1}},
\\
 \textbf{Jiachen T. Wang\textsuperscript{2}},
 \textbf{Jian Du\textsuperscript{1}},
\\
\\
 \textsuperscript{1}TikTok Inc.,
 \textsuperscript{2}Princeton University,
\\
 \small{
 }
}

\usepackage{hyperref}       
\usepackage{url}            
\usepackage{booktabs}       
\usepackage{amsfonts}       
\usepackage{nicefrac}       
\usepackage{microtype}      
\usepackage{xcolor}         

\usepackage{comment}
\usepackage{amsmath}

\usepackage{epsfig}
\usepackage{graphicx}
\usepackage{wrapfig}
\usepackage{subfigure}
\usepackage{multirow}
\usepackage{hyperref}
\usepackage{graphicx}
\usepackage{caption}
\usepackage{array}
\usepackage{bbm}
\usepackage{color}
\usepackage{enumerate}
\usepackage{enumitem}
\setlist{leftmargin=10mm}
\usepackage{mathtools}
\usepackage{amsmath,amssymb,amsthm,bm}
\usepackage{amsfonts,graphicx}
\usepackage{mathrsfs}
\usepackage{algorithm}
\usepackage{algpseudocode}

\usepackage{listings}
\usepackage{lipsum} 
\usepackage{tcolorbox}
\usepackage{cuted}  
\usepackage[normalem]{ulem} 
\usepackage{soul}           
\newif\iffinal

\iffinal
    \newcommand{\tianhao}[1]{}
    \newcommand{\jiandu}[1]{}
    \newcommand{\yq}[1]{}

\else
    \newcommand{\tianhao}[1]{{\bf \textcolor{purple}{[Tianhao: #1]}}}
    \newcommand{\jiandu}[1]{{\bf \textcolor{blue}{[jiandu:#1]}}}
    \newcommand{\yq}[1]{{\bf \textcolor{blue}{[Yuqing:#1]}}}

\fi

\usepackage{cleveref}
\usepackage{bbm}

\newtheorem{theorem}{Theorem}
\newtheorem{lemma}[theorem]{Lemma}

\newtheorem{definition}[theorem]{Definition}

\newtheorem{remark-star}{Remark}
\newtheorem{remark-star-1}{Remark}

\newtheorem*{proof-sketch}{Proof Sketch}
\usepackage{booktabs}

\DeclareMathOperator*{\argmax}{\arg\!\max}

\newcommand{\K}{\mathbb{K}}

\newcommand{\V}{\mathbb{V}}

\newcommand{\prob}{\mathbb{P}}

\newcommand{\M}{\mathcal{M}}

\newcommand{\U}{v}

\newcommand{\xsj}{x^{(S)}_{j}}

\newcommand{\zval}{ \hat{z} }
\newcommand{\xval}{ \hat{x} }
\newcommand{\yval}{ \hat{y} }
\newcommand{\ks}{K^{(S)}}

\newcommand{\shapvalue}{\text{TokenShapley}~}
\newcommand{\concite}{\text{ConCite}~}
\newcommand{\hidsim}{\text{HidSim}~}
\newcommand{\embsim}{\text{EmbSim}~}

\begin{document}
\maketitle
\begin{abstract}
Large language models (LLMs) demonstrate strong capabilities in in-context learning, but verifying the correctness of their generated responses remains a challenge. Prior work has explored attribution at the sentence level, but these methods fall short when users seek attribution for specific keywords within the response, such as numbers, years, or names. To address this limitation, we propose TokenShapley, a novel token-level attribution method that combines Shapley value-based data attribution with KNN-based retrieval techniques inspired by recent advances in KNN-augmented LLMs. By leveraging a precomputed datastore for contextual retrieval and computing Shapley values to quantify token importance, TokenShapley provides a fine-grained data attribution approach. Extensive evaluations on four benchmarks show that TokenShapley outperforms state-of-the-art baselines in token-level attribution, achieving a 11–23\% improvement in accuracy.
\end{abstract}


\newcommand{\reza}[1]{{\textcolor{red}{RS: #1}}}

\section{Introduction}

In context learning (ICL) \citep{brown2020language, dong2024survey} has emerged as a powerful capability of large language models (LLMs) \citep{achiam2023gpt, dubey2024llama, guo2025deepseek}. LLMs can perform a wide range of tasks (for example, question answering \citep{choi2018quac}, solving math problems) by learning from the context information users provide. ICL is particularly valuable in scenarios where data collection and retraining are costly or impractical (for example, medical diagnosis \citep{wang2023chatcad}).

\begin{figure}[h] 
    \centering
    \includegraphics[width=0.48\textwidth]{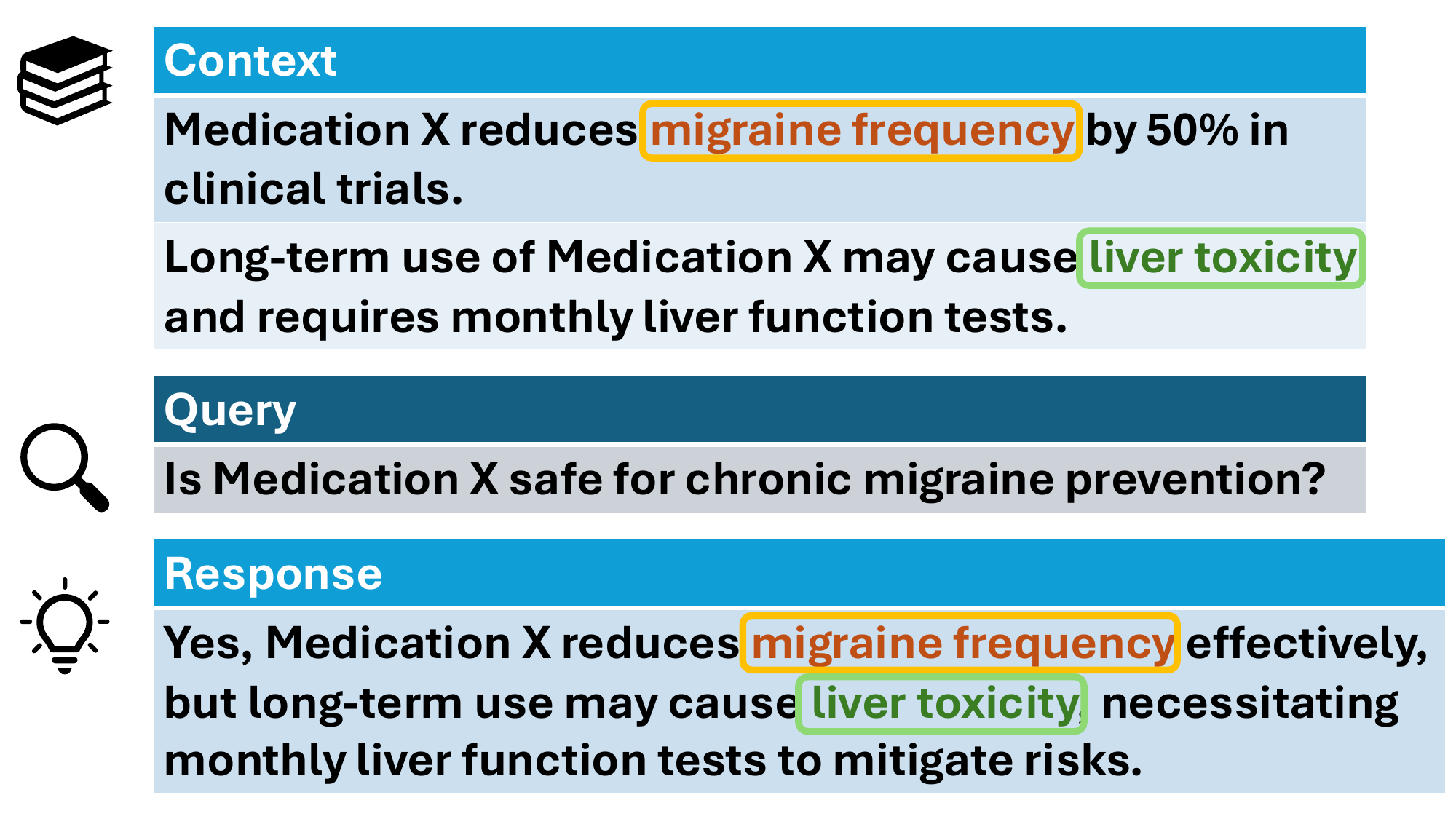}
    \caption{An example of context attribution. Here the response contains multiple information. Doing sentence-level attribution may provide incomplete evidence. A better way is to provide token-level attribution.}
    \label{fig:attribution}
\end{figure}

However, responses generated by large language models (LLMs) are not inherently grounded in truth, as these models can sometimes hallucinate \citep{zhang2023siren}, producing factually incorrect or misleading information. This can lead to serious risks, particularly in high-stakes applications (e.g., healthcare diagnosis). As such, fact-checking \citep{quelle2024perils} and context attribution \citep{li2023survey} is crucial to ensure that LLM responses are both interpretable and verifiable. Consider an LLM that recommends medication based on a medical database (Figure \ref{fig:attribution}). A user may want to get evidence from key tokens, such as "migraine frequency" and "liver toxicity", that capture comprehensive evidence of both benefits and risks, which might be overlooked with a single attributed sentence in the context. By pinpointing the tokens behind a response and providing supporting evidence, users can more easily assess its reliability.

Approaches to attribution in LLMs generally fall into three categories \citep{li2023survey}. 1) \textbf{Direct Generated Attribution} \citep{sun2022recitation, weller2023according}. These methods prompt LLMs to generate self-attributions to improve response accuracy. However, researchers have found that LLMs often struggle to provide reliable evidence, particularly for domain-specific questions \citep{gravel2023learning}.  2) \textbf{Post-Retrieval Answering} \citep{asai2023self, li2023llatrieval}. These approaches involve additional model updates or training to improve response accuracy. However, the requirement for direct access to the model’s parameters can introduce additional computational overhead. 3) \textbf{Post-Generation Attribution} \citep{gao2022rarr, huo2023retrieving, chen2023complex, cohen2024contextcite}. These methods utilize search engines, document retrieval systems, or surrogate models to retrieve evidence based on input questions and generated answers. They primarily attribute information at the sentence level, which lacks fine-grained attribution when a sentence contains multiple pieces of information. See Figure \ref{fig:attribution} for an example.

To address the limitations of existing attribution methods, we aim to propose a new attribution method that 1) Provides accurate evidence. 2) Eliminates the need for additional tuning with LLMs. 3) Enables fine-grained attribution. This motivates us to adopt the Shapley value \citep{shapley1953value}, which fairly attributes contributions by distributing credit among input components based on their marginal impact. Shapley values have been widely used for training data attribution \citep{jia2019efficient, jia2019towards, wang2024efficient}. In supervised learning tasks, when a K-nearest neighbors (KNN) model is used for attribution, Shapley values can be computed in polynomial time \citep{wang2024efficient}. However, applying KNN for attribution in LLMs remains challenging due to the vague notion of "labels" among tokens.

\begin{figure*}[h] 
    \centering
    \includegraphics[trim=10 130 10 130, clip, width=\linewidth]{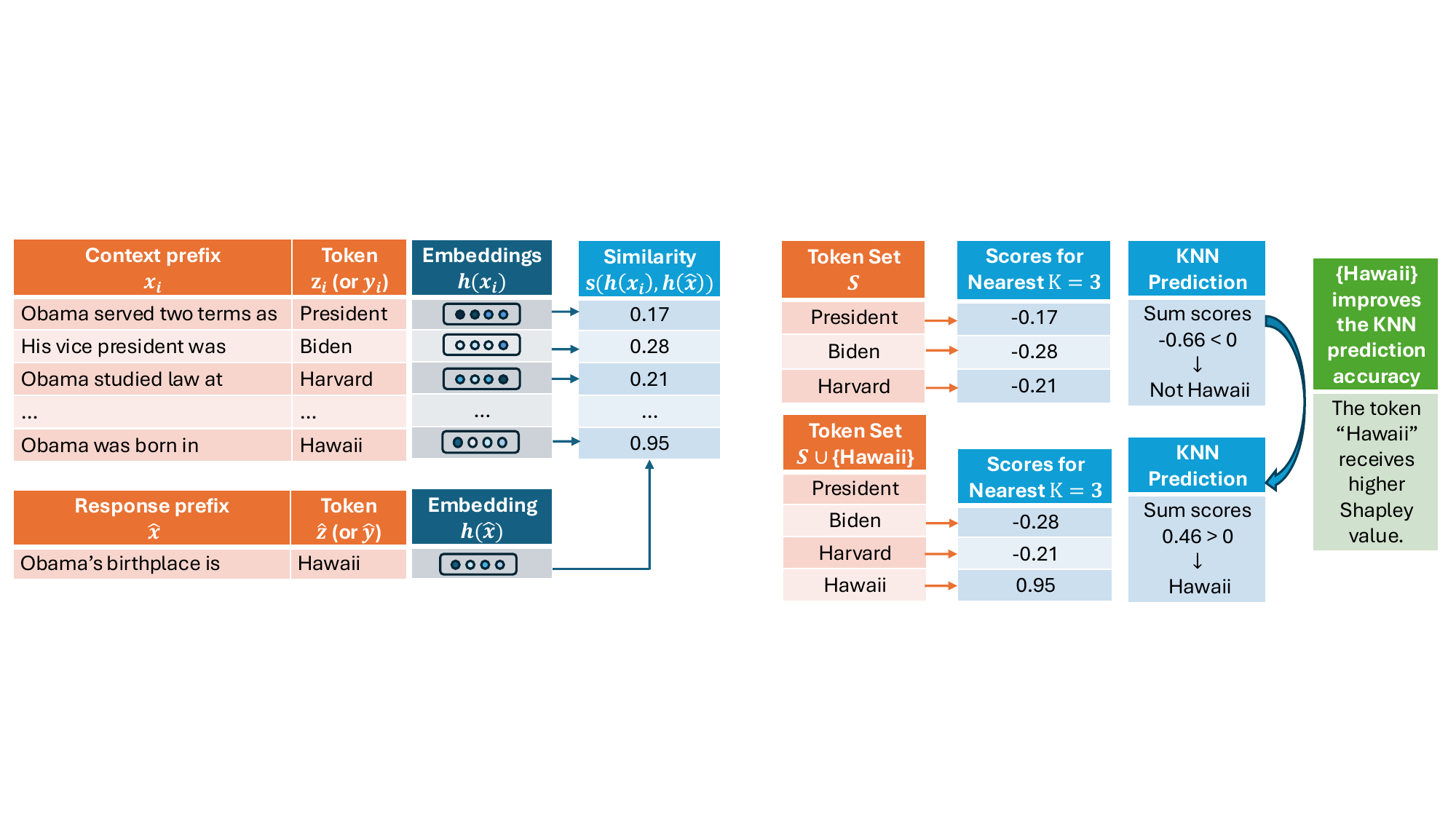}
    \caption{An illustration of TokenShapley. Given a response token $\hat{z}$, we first compute similarity scores using the datastore. To determine the Shapley value of a context token $z_i$, we select all possible subsets $S \subseteq D \backslash \{z_i\}$. We then run a KNN classifier on both $S$ and $S \cup \{z_i\}$ to evaluate whether adding $z_i$ improves the accuracy of predicting $\hat{z}$. If the accuracy increases, $z_i$ is assigned a higher Shapley value.}
    \label{fig:flow-chart}
\end{figure*}

We observe that KNN-LM \citep{k2019knn, li2024nearestneighborspeculativedecoding} enhances LLM generation by leveraging a KNN model, which stores (key, value) pairs for retrieval, where the key represents the prefix of a given token and the value is the token itself. Building on this idea, we introduce TokenShapley, a novel token-level attribution method that employs a KNN model specifically for attribution. The flowchart of \shapvalue is shown in Figure \ref{fig:flow-chart}, with further details provided in Section \ref{sec:algo}. 

In summary, our contributions are
\begin{itemize}
    \item We propose a token-level attribution algorithm TokenShapley, allowing accurate, efficient and fine-grained attribution.
    
    \item We introduce an efficient strategy for exact token-level Shapley value calculation that leverages embedding-based retrieval and weighted KNN search, eliminating the need for Monte Carlo sampling.
    
    \item TokenShapley improves token-level attribution accuracy by 11–23\% on Verifiability-Granular and QuoteSum, achieves perfect score on KV Retrieval, and raises Natural Questions precision by about 3.2\%.
\end{itemize}

\section{Related work}


We provide additional related work not discussed in the previous section.

\noindent\textbf{External Data Attribution} Many prior work focus on tracing model's prediction back to the training data, this includes methods such as influence functions~\citep{koh2017understanding, hammoudeh2022identifying, grosse2023studying, mlodozeniec2024influence}. \cite{slobodkin2024attribute, sancheti2024post} provides citations to sentences in the response from external documents. 


\noindent\textbf{Context Attribution} 
\citep{cohen2024contextcite} learns a linear surrogate model that approximates the external context's behavior towards the generated response. They partition the context into sentences and learn a score of each sentence towards the response. \cite{park2023trak} propose the linear data modeling score, which measures the extent to which attribution scores can predict the effect of excluding a random subset of sources. \citep{phukan2024peering, ding2024attention, qi2024model} utilize hidden-space embeddings of tokens in both the context and response to compute similarity scores. The tokens in the context with the highest scores are then identified as attributions for the corresponding tokens in the response.




\section{Problem Setup and Background}

In this section, we will start with the problem setup of context attribution. The notation used in this paper is listed in Table \ref{tab:notation} in Appendix \ref{app:weight-knn} .

\noindent\textbf{Problem Setup} We consider the in-context learning scenario in which a language model generates a response to a query based on a given context, such as in search engines like Bing Chat, which answers users' questions using information from news webpages. After seeing the response, a user may wonder: Is the generated response correct? Which part of the context led the model to produce this response? Our goal is to pinpoint the specific tokens from the context that drive the model's response, thereby enabling precise token-level attribution. 

Formally, given a query $q$ and a corpus of text data (context) with $N$ tokens, $D = (z_1, \cdots, z_N)$, the language model generates a response $R$ with $T$ tokens $R=(\hat{z}_1, ..., \hat{z}_T)$. For each response token $\hat{z} \in R$, we aim to identify the most contributive tokens $z \in D$. To achieve this, we compute a score $\phi_{z \to \hat{z}}$ to quantify the contribution of $z$ to $\hat{z}$. Tokens $z$ with the highest scores $\phi_{z \to \hat{z}}$ are selected as the attribution for the token $\hat{z}$. We simply write $\phi_z$ when $\hat{z}$ is clear from the context. A widely used scoring function is the Shapley Value \citep{shapley1953value}, which we will introduce in the following section.

\subsection{Shapley Value}

The Shapley value (SV) \citep{shapley1953value} is a classic concept from game theory to attribute the total gains generated by the coalition of all players. Formally, given a utility function $\U(\cdot)$ (e.g., the model's prediction accuracy) and a dataset $D$, the Shapley value assigns a score to each data point $z \in D$ representing its contribution of $z$ to $\phi_z(\U)$.

To compute this contribution, one utility function is evaluated with the data point $z$ present and another utility function with $z$ excluded, denoted as $\U(S\cup {z}) - \U(S)$, where $S$ is any possible subset of the dataset. The difference in utility functions represents the marginal contribution of $z$. The Shapley value is then the weighted average of these differences across all possible subsets $S$.

\begin{definition} \label{def:shapley}
Given a utility function $v(\cdot)$ and a dataset $D$ with $N$ data points, the Shapley value of a data point $z \in D$ is defined as 
\begin{align}
\label{eq:def-shapley}
    \phi_z\left( \U \right) = \frac{1}{N} \sum_{k=1}^{N}  \sum_{\substack{S \subseteq D_{-z} \\ |S|=k-1}} \frac{ \U(S \cup \{z\}) - \U(S)}{{N-1 \choose k-1}}
\end{align}
Here $D_{-z} = D \backslash  \{z\}$. When the utility function $\U(\cdot)$ is clear from the context, we use $\phi_z$ as a shorthand to represent the Shapley value of data point $z$.
\end{definition}

Equation \ref{eq:def-shapley} shows that computing the exact Shapley value for a data point $z$ requires evaluating the utility function $\U(\cdot)$ for all possible subsets $S \subseteq D \backslash \{ z\}$. The number of evaluations grows exponentially with respect to $N$ (the size of $D$), making the exact calculation computationally hard.

Fortunately, recent research \cite{ghorbani2022data, liang2020beyond, liang2021herald, courtnage2021shapley} has shown that the exact Shapley value for $K$-nearest neighbors (KNN)-based machine learning models can be computed with polynomial-time complexity.


\subsection{KNN-Shapley}\label{sec: shapley}

KNN-Shapley \citep{jia2019towards} was originally introduced to address data attribution in the training set, which uses KNN to retrieve training samples. KNN-Shapley aims to determine the contribution of each training data point to the model prediction. In this section, we provide background information and highlight the challenges of applying KNN-Shapley to context attribution.

We use a $K$-nearest neighbors classifier to illustrate how KNN-Shapley works. Given a validation data point $\zval = (\xval, \yval)$, where $\xval$ is the feature and $\yval$ is the label, a KNN classifier predicts $\yval$ via a weighted majority vote of the top-$K$ neighbor's labels from the training set $D$. The weight assigned to a data point $z_i = (x_i, y_i) \in D$  is based on the similarity metric $s(\cdot, \cdot)$ between the feature $x_i$ and the validation point feature $\xval$.

\citep{jia2019towards, wang2023noteknn} defines the utility function as the likelihood of predicting the correct label $\yval$. Let $x^{(S)}_j$ refer to $j$-th nearest feature in the set $S$ with respect to $\xval$, and $y^{(S)}_j$ be the label for $x^{(S)}_j$. Let $\ks=\min(K, |S|)$ be the number of voting points (in case $S$ has less than K points). The utility function can be expressed as:
 
\begin{align}
    \U(S; \zval) 
    =\frac{\sum_{j=1}^{\ks} s( \xsj, \xval)\mathbb{I}[y_{j}^{(S)}=\yval]}{\sum_{j=1}^{\ks}s(\xsj, \xval)}
    \label{softknn}
\end{align}

When the weights $s(\xval, \xsj)$ are 1, the utility function gives the portion of voting points that correctly predict the label $\hat{y}$. The utility function of $S$ only depends on the top-$K$ neighbors that are within $S$. 



For weighted KNN, \citep{wang2024efficient} demonstrates that by discretizing and slightly tweaking the utility function and using dynamic programming, it is possible to compute the exact Shapley values in $O(N^2)$ runtime. The utility function for weighted KNN is written as 
\begin{align}
\label{eq:hard-label}
 &\U(S; \zval) \\
 \nonumber
 =& \mathbb{I} \bigg[ \yval \in \argmax_{w\in \mathcal{C}}
    \sum_{j=1}^{\ks} 
    s(\xsj, \xval) \mathbb{I}[y_j^{(S)}=w]\bigg]
\end{align}
Here $\mathcal{C}$ represents the space of classes, the number of labels is $|\mathcal{C}|$.  This utility function measures whether the weighted majority vote predicts the true label $\hat{y}$. Compared to the equation~\ref{softknn}, this utility function omits the normalization term. 


The limitation of the KNN-Shapley method is that the utility function is primarily designed for supervised learning tasks. It relies on labeled data, where the correctness of a prediction serves as the utility function. However, in large language model generation task, defining an appropriate utility function remains an open challenge \citep{jia2019efficient, wang2024efficient}, making it unclear how to compute an exact Shapley value in polynomial time. One of our key contributions is introducing a novel approach to creating "labels" within the generation process. Before presenting the details in Section \ref{sec:algo}, we first provide background on the KNN Language Model \citep{k2019knn}, which gives the intuition behind our design.

\subsection{KNN-LM}
The primary challenge of applying KNN-Shapley in language models (LMs) is the absence of an integrated KNN mechanism. Can we leverage such mechanism to enhance model's generation capabilities? KNN-LM \citep{k2019knn, li2024nearestneighborspeculativedecoding} offers a viable solution. It builds a KNN model on a datastore constructed from a text corpus and utilizes it to improve model performance. 

Let $x_t = (z_1, \cdots, z_{t-1})$ be the prefix tokens and $y_t=z_t$ be the next token. The LM generates $y_t$ based on the probability distribution $p(y_t | x_t)$. The KNN-LM enhances a pre-trained language model with a KNN retrieval mechanism, which processes a text collection in a single forward pass. The resulting context-target pairs are stored in a key-value datastore. During the inference, the datastore is queried to provide additional information to the LMs. Next we describe how this datastore is constructed.



\subsubsection{Datastore}

Let $h(\cdot)$ be the hidden state representation function that maps the prefix tokens $x_t = (z_1, \cdots, z_{t-1})$ to a LM hidden layer representation. For a token in the text corpus $y_t = z_t \in D$ and its prefix $x_t = (z_1, \cdots, z_{t-1})$, the (key, value) pair $(k_t, v_t)$ is set such that the key $k_t $ is the hidden layer representation $ h(x_t)$ and the value $v_t $ is the target token $y_t$. The datastore consists of all key-value pairs generated from the dataset $D$.


\begin{align}
    (\mathcal{K}, \mathcal{V}) = \{(h(x_t), y_t) | z_t \in D \}
\end{align}

\subsubsection{Inference}

During inference, given the prefix tokens $\hat{x}$, the model generates the target token according to $p_{LM}(\hat{y} | \hat{x})$. It also gives the hidden state representation $h(\hat{x})$. The model queries the datastore with $h(\hat{x})$ to retrieve its $K$-nearest neighbors $\mathcal{N}$ according to a similarity function $s(\cdot, \cdot)$.


\begin{align}
\label{eq:knn-prob}
    p_{K}(\hat{y} | \hat{x}) \propto \sum_{\substack{(h(x_t), y_t) \\ \in \mathcal{N}}}  s(h(x_t), h(\hat{x})) \mathbb{I}[y_t = \hat{y}]
\end{align}

The similarity function used is the RBF kernel function, $s(h, h') = \exp (- \gamma \|h - h' \|_2^2)$, where $\gamma > 0 $ is a parameter and  $h, h'$ are two vectors. The final KNN-LM distribution is a weighted combination of the nearest neighbor distribution $p_{KNN}$ and the model distribution $p_{LLM}$ with a parameter $0\leq \lambda \leq 1 $, $    p(y | x) = \lambda p_{K}(y | x) + (1 - \lambda) p_{LM}(y | x)$.

We observe that the probability distribution in Equation \ref{eq:knn-prob} is closely related to the utility function in Equations \ref{eq:hard-label}, as they both measure the likelihood of predicting the label $\hat{y}$ with the KNN model. This motivates us to create a KNN model in the LM generation. In this setting, we say the (feature, label) pair for a data point $z_t \in D$ is a (prefix, token) pair which is represented by $ (x_t, y_t)$. We can then apply the weighted KNN-Shapley method using a utility function similar to Equation \ref{eq:hard-label}. We provide the details of the algorithm in the next section.

\section{Algorithm}
\label{sec:algo}



The goal of context attribution is to identify the most contributive sources within the context $D$ that affect the model's response $R$. Specifically, we aim to assign scores to each token in the context $z_i \in D$, reflecting its contribution to a given token in the response $\hat{z}_t \in R$. 

Recall that, in order to run the KNN-Shapley value algorithm which gives exact Shapley values in polynomial run time, we need to design a KNN model in language model generation. In Section \ref{sec:datastore} we follow the design of KNN-LM to create a datastore with (key, value) pairs for the KNN model. Then in Section \ref{sec:utility}, we design the utility function $\U(\cdot)$ for efficient KNN-Shapley value. In Section \ref{sec:weight}, we briefly review the core technique used in \citep{wang2024efficient}. Lastly, we show how to accumulate Shapley values for attribution from tokens to tokens. Algorithm \ref{alg:shapley} shows the pseudo-code for the proposed \shapvalue algorithm.

\begin{algorithm}
\caption{TokenShapley}\label{alg:shapley}
\begin{algorithmic}[1]
\Require Key-value data store cache $\mathcal{C}$. Embedding model $h(\cdot)$. Context $D=(z_1, \cdots, z_m)$. Query $q$. Response $R=(\hat{z}_1, \cdots, \hat{z}_T)$. Token $\hat{z}_t \in R$ of interest.
\Ensure The Shapley value of each token $z_i$ in context $D$ w.r.t. token $\hat{z}_t$: $\phi_{{z_i} \to \hat{z}_t}$.
\State Get the "feature" $\hat{x}_t \gets q + R[1:t]$.
\State Get the "label" $\hat{y}_t \gets \hat{z}_t$.
\State Get the hidden representation $h(\hat{x}_t)$.
\State Search in the data store cache $\mathcal{C}$ using $h(\hat{x}_t)$ as the key to find the nearest $K$ neighbors $(x_1, \cdots, x_K)$.
\State Calculate the utility function $\U(\cdot)$ according to Equation \ref{eq:knnscore}.
\State Run Weighted KNN Shapley algorithm to get the Shapley value for $z_i$ (w.r.t. token $\hat{z}_t$).
\State
\Return $\{\phi_{{z_i} \to \hat{z}_t}\}_{i=1}^{N}$.
\end{algorithmic}
\end{algorithm}

\subsection{Build Datastore}
\label{sec:datastore}

Suppose there are $N$ tokens in the context $D = (z_1, \cdots, z_N)$. For the t-th token $z_{t}$, the prefix of $z_t$ is $x_{t}=(z_{t_0}, z_{t_0 + 1}, \cdots, z_{t-1})$. Here $t_0$ is the starting index for the prefix, for example, we can set $t_0$ as the index for the starting token at the current sentence of $z_t$. In this setting, we say a data point $z_t \in D$ has a (feature, label) pair which is represented by the (prefix, token) pair $z_t \to (x_t, y_t)$.

We use the encoder $h(\cdot)$ of the embedding model to encode each prefix sentence $x_t$ into a hidden states $h(x_t)$. Let $y_t$ be the next token following $x_t$ in the context $D$. Given a token $z_{t}\in D$, we define the key-value pairs in datastore as below
\begin{equation}
    (\mathcal{K}, \mathcal{V})= \{(h(x_{t}), y_{t})|z_{t} \in D \}
\end{equation}
The size of the datastore is proportional to the total number of tokens within $D$.

\subsection{Define Utility Function} 
\label{sec:utility}

During inference, the language model outputs the next token distribution $f_{LM}(\hat{y}_t|\hat{x}_{t})$ along with the hidden state $h(\hat{x}_t)$. We use $h(\hat{x}_t)$ as a query to search the datastore $(\mathcal{K}, \mathcal{V})$ and retrieve the $K$-nearest neighbors based on the distance in the hidden space. Next, we apply weighted KNN-Shapley \citep{wang2024efficient} (see Sec \ref{sec: shapley}) to calculate Shapley value for selected neighbors. This method relies on evaluating the  utility function of all possible subset $S$ within $(\mathcal{K}, \mathcal{V})$. 



\begin{align}
\label{eq:knnscore}
&\U(S; \hat{z}_t)\\
\nonumber
=&\begin{cases}
1, & \sum_{j=1}^{\ks} s(h(\hat{x}_t), h(\xsj))\mathbb{I}[y^{(S)}_{j}=\hat{y}_t] \\
& \geq \sum_{j=1}^{\ks} s(h(\hat{x}_t), h(\xsj))\mathbb{I}[y^{(S)}_{j} \neq \hat{y}_t] \\
0, &  otherwise.
\end{cases}
\end{align}

The utility function evaluates whether the KNN predictor correctly predicts the generated token $\hat{y}_t$. If the majority vote says the label should be $\hat{y}$, the utility function will be 1. If the majority vote favors the label (not $\hat{y}$), the utility function will be 0. It's a special case of Equation \ref{eq:hard-label}, here it's a binary classification problem. The similarity function used is the RBF kernel function, $s(h, h') = \exp (- \gamma \|h - h' \|_2^2)$, where $\gamma > 0 $ is a parameter and  $h, h'$ are two vectors. As a special case, when $K=1$, the utility function is 
\begin{align}
\label{eq:k=1}
    \U(S; \hat{z}_t) = \mathbb{I}[y^{S}_{1}=\hat{y}_t]
\end{align}

The utility function \ref{eq:k=1} gives value 1 if and only if the nearest data point to $\hat{z}_t \to (\hat{x}_t, \hat{y}_t)$ is $z_1^{(S)}  \to (x_1^{(S)}, y_1^{(S)}) $ and the token $y_1^{(S)}$ is exactly the same as $\hat{y}_t$. Note that we only need to calculate the KNN once for each label $\hat{y}_t$, if $S$ contains no data points selected by the KNN, the utility function $\U(S;\hat{z}_t)$ will always be 0, and we do not need to update it.

\subsection{Calculate Shapley Values}
\label{sec:weight}

In this section, we provide a brief description of how to calculate the Shapley value using the algorithm presented in \citep{wang2024efficient} to make the discussion self-contained. 

\begin{lemma}\citep{wang2024efficient}
For any data point $z_i \in D$ and any subset $S \subseteq D \setminus \{z_i\}$, the marginal contribution has the expression as follows:
\begin{align}
\label{eq:diff}
    &v(S \cup \{z_i\}) - v(S) \\
    \nonumber
    =& \begin{cases}
        1 & \text{if } y_i = \yval, \text{Cond}_{KNN}, \text{Cond}_{0\to1} \\
        -1 & \text{if } y_i \neq \yval, \text{Cond}_{KNN}, \text{Cond}_{1\to0} \\
        0 & \text{Otherwise}
    \end{cases}
\end{align}
\end{lemma}

Here, $\text{Cond}_{KNN}$ is the condition when $z_i$ is within the $K$ nearest neighbors of $\zval$ among $S \cup {z_i}$. If $z_i$ is not one of the neighbors, adding $z_i$ to the set $S$ will have no effect on the utility function, and the difference will be 0. $\text{Cond}_{0 \to 1}$ is the condition when adding $z_i$ changes the majority vote in Equation \ref{eq:knnscore} from a label (not $\hat{y}$) to $\hat{y}$. This can only happen when $z_i$'s token $y_i$ is the same as the label, i.e., $y_i = \hat{y}$. Similarly, $\text{Cond}_{1 \to 0}$ is the condition when adding $z_i$ changes the majority vote in Equation \ref{eq:knnscore} from label $\hat{y}$ to a different label (not $\hat{y}$). This can only happen when $z_i$'s token differs from the label, i.e., $y_i \neq \hat{y}$. The exact expressions for these conditions are provided in Appendix \ref{app:weight-knn}.

Given the expression in Equation \ref{eq:diff}, the Shapley value calculation can be efficiently computed using the following theorem.

\begin{theorem}
\citep{wang2024efficient}
    Let $\mathcal{G}_{i,\ell}$ denote the count of subsets $S \subseteq D \setminus \{z_i \}$ of size $\ell$ that satisfy \textit{(1)} $\textit{Cond}_{KNN}$, and \textit{(2)} $\textit{Cond}_{0\to1}$ if $y_i = \yval$, or $\textit{Cond}_{1\to0}$ if $y_i \neq \yval$. Then for a weighted KNN binary classifier using the utility function $\U(S; \hat{x})$ in Equation \ref{eq:knnscore}, the Shapley value of a data point $z_i$ can be expressed as:
\begin{equation}
\phi_{z_i \to \hat{z}}(\U) = \frac{2\mathbb{I}[y_i = \yval] - 1}{N} \sum_{\ell=0}^{N-1} \frac{\mathcal{G}_{i,\ell}}{\binom{N-1}{\ell} }
\end{equation}
The computation of all $\mathcal{G}_{i,\ell} $ can be done in $O(N^2)$ time complexity, and the time complexity for calculating all Shapley values is $O(N^2)$. 
\end{theorem}


The computation of all $\mathcal{G}_{i,\ell}$ is a counting problem that can be efficiently solved using dynamic programming. From Equation \ref{eq:diff} and $\text{Cond}_{KNN}$, we observe that if $z_i$ is not among the $K$ nearest neighbors, its contribution is zero. Therefore, we can further enhance efficiency by selecting only the $M$ nearest neighbors and updating the Shapley value exclusively for the data points within these neighbors ($K < M < N$).


\subsection{Accumulate Attribution Scores} 
\label{sec:sum}

In the previous sections, we demonstrated how to compute the Shapley value for attributing from a single context token $z_i$ to a response token $\hat{z}$. A natural extension is to determine the attribution from a set of context tokens $z_i \in S$ (e.g., a sentence in the context) to a set of response tokens $\hat{z} \in \hat{S}$ (e.g., the entire response). This can be achieved by summing the individual Shapley value scores.

\noindent\textbf{Step 1} Recall that $\phi_{z_i \to \zval}(\U)$ measures the contribution of a data point $z_i \in D$ for a specific token $\zval$. To compute the Shapley value over the token set $\hat{S}$ of interest, we leverage the linearity property of the Shapley value. Specifically, this is done by summing the individual Shapley values for each validation point, i.e., $\phi_{z_i \to \hat{S}}(\U):= \sum_{\zval\in \hat{S}} \phi_{z_i \to \hat{z}}(\U)$.

\noindent\textbf{Step 2} We sum up the Shapley values in the context token sets $S$ to get the attribution from $S$ to $\hat{S}$, i.e., $\phi_{S \to \hat{S}}(\U):= \sum_{z_i \in S} \phi_{z_i \to \hat{S}}(\U)$. This leads to the final expression $\phi_{S \to \hat{S}}(\U):= \sum_{z_i \in S} \sum_{\zval\in \hat{S}} \phi_{z_i \to \hat{z}}(\U)$.




\section{Experiments}
\label{sec:eval}

In this section, we compare TokenShapley with several baselines across four datasets. We use a single A100 GPU with 80 GB memory for experiments. For parameters in TokenShapley, we set $K=1$ and $M=10$ in all experiments.


\subsection{Baselines}
We consider the following baselines on model explanations and attributions. 

\noindent\textbf{Hidden Space Similarity} (HidSim for short) \citep{phukan2024peering} measures similarity by comparing hidden space embeddings of tokens in the context and response. For each token in the response, the most similar tokens in the context are identified as its attribution.

\noindent\textbf{Text Embedding Similarity} (EmbSim for short): We use a "text-embedding-3-small" model from OpenAI to generate embeddings for each source and the generated response. The cosine similarity between the source embedding and the response embedding is used as the attribution score for the source.

\noindent\textbf{ContextCite} (ConCite for short) \citep{cohen2024contextcite} trains a linear surrogate model to trace model responses back to their contexts. It assigns a weight to each source (i.e., sentence) as an attribution score, where higher scores indicate greater contributions to the model's response. ConCite is primarily designed for sentence-level attribution, as training a linear surrogate model for each source is computationally expensive for token-level attribution.


\subsection{Open Source LLMs}
We list the three open-source large language models used in the experiment. When the context is clear, we omit model parameters—for instance, 'Llama-3' refers to Llama-3-8B as used in the experiments. 1) \textbf{Llama-3-8B}. The Hugging Face model name is "meta-llama/Meta-Llama-3.1-8B-Instruct". 2) \textbf{Mistral-7B}. The Hugging Face model name is "mistralai/Mistral-7B-Instruct-v0.3". 3) \textbf{Yi-6B}. The Hugging Face model name is "01-ai/Yi-6B-Chat".

\subsection{Datasets}
We use five datasets for evaluation.

\noindent\textbf{QuoteSum} \citep{schuster2023semqa}. The context comprises multiple passages, with the answer to the question provided. Several key words are manually marked and labeled with the index of the passage from which their attribution originates.


\noindent\textbf{Verifiability-Granular} \citep{phukan2024peering}. The context consists of a list of passages, along with a given question and answer. Several key tokens are manually marked, and the task is to identify the passage most relevant to these tokens. The ground truth passage is also manually labeled.

\noindent\textbf{KV Retrieval} \citep{liu-etal-2024-lost}. The context is a sequence of (key, value) (randomly generated), and the query is to find the value corresponding to a key, the task is to just search and recall information from the context.

\noindent\textbf{Natural Questions} \citep{kwiatkowski2019natural}. In this dataset, the context is a Wikipedia page, a question is asked and a labeled answer is given. We use the labeled answer (original text from the Wiki page) as the true attribution source. We select answers that consist solely of text (marked as paragraphs in the HTML tags).


\noindent \textbf{CNN Dailymail} \citep{nallapati2016abstractive}. The Dataset is an English-language dataset containing news articles as written by journalists at CNN and the Daily Mail. We prompt the language model
to briefly summarize the news article.

\begin{figure*}[h!] 
\centering\includegraphics[width=\textwidth]{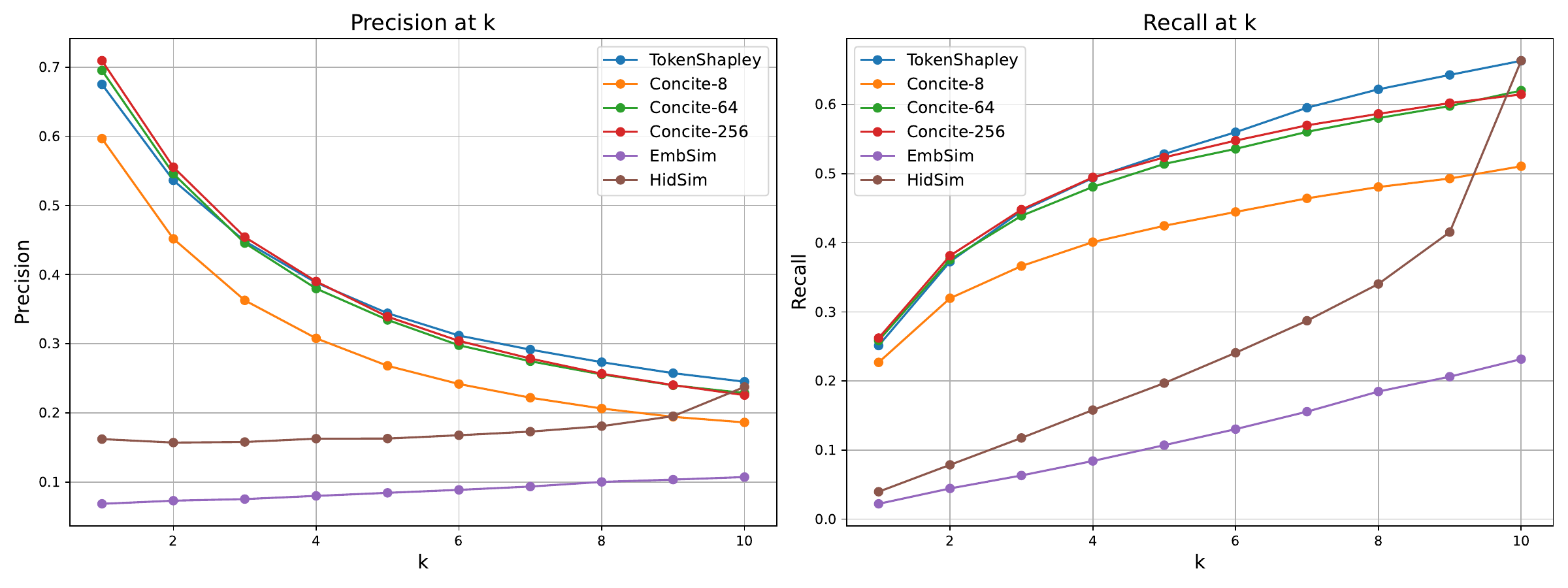}
\caption{Precision and recall of top-k attributed sentences compared with ground truth attribution. Precision is defined as (number of match) / (k), recall is defined as (num of match) / (number of ground truth sentences).}
    \label{fig:nq}
\end{figure*}

\subsection{Metric}


For the first three datasets, we use prediction accuracy as the evaluation metric, since ground-truth attribution labels are available. For the Natural Questions dataset, we report precision, recall, and F1 score. For the CNN/DailyMail dataset, we adopt the log probability drop metric \citep{cohen2024contextcite}, which measures the decrease in log probability of generating the same response when the attributed sentences are removed from the context.

\subsection{Experiment on QuoteSum}

\begin{table}[h!]
\centering
\begin{tabular}{lccc}
\toprule
\textbf{QuoteSum} & Llama-3 & Mistral & Yi \\
\midrule
\hidsim$^{*}$   & - & 89.95 &89.24\\
\hidsim & 83.91 & 80.73 & 75.59\\
\shapvalue & \textbf{92.51} &\textbf{92.20}  & \textbf{91.27}\\
\midrule
\multicolumn{3}{l}{\embsim with OpenAI text embedding} & 53.38 \\
\bottomrule
\end{tabular}
\caption{Attribution Accuracy on QuoteSum dataset with 1319 examples. \hidsim$^{*}$ quoted the results from \citep{phukan2024peering}.}
\label{tab:qsum}
\end{table}


In our experiments on the QuoteSum dataset, attribution accuracy varied noticeably across the baselines. Our TokenShapley method consistently outperformed the others, scoring over 91\% on all models. TokenShapley works well with QuoteSum because the dataset is designed with clearly marked short answers in the summaries, where each answer is tied to a specific passage via bracketed indices. Our method focuses on individual tokens that form these short answers, and results in higher performance. We also observe that HidSim suffer from slight misalignments in token boundaries and hidden state similarities, resulting in lower accuracy. EmbSim’s poorer performance arises from the differences in embedding spaces between the response and source, which makes it less sensitive to detailed token-level information. 



\subsection{Experiment on Verifiability-Granular}

\begin{table}[h!]
\centering
\begin{tabular}{lccc}
\toprule
\textbf{VeriGran} & Llama-3 & Mistral & Yi \\
\midrule
\hidsim$^{*}$ & - & 77.71 & 77.61 \\
\hidsim & 70.77 & 71.28 & 61.54\\
\shapvalue & \textbf{82.23} & \textbf{84.77} &\textbf{84.77}\\
\midrule
\multicolumn{3}{l}{\embsim with OpenAI text embedding} & 70.05 \\
\bottomrule
\end{tabular}
\caption{Attribution Accuracy on Verifiability-Granular dataset with 197 examples. \hidsim$^{*}$ quoted the results from \citep{phukan2024peering}.}
\label{tab:verify}
\end{table}

The results in Table \ref{tab:verify} show that TokenShapley outperforms both HidSim and EmbSim across all models. TokenShapley achieves the highest accuracy, with improvements of 4.5\% over HidSim and 10.7\% over EmbSim for Llama-3, and similarly for other models. This shows that token-level attribution is more effective for this dataset, as it better captures fine-grained relationships between the response and passages. On the other hand, EmbSim's sentence-level approach is less sensitive to the detailed contextual relevance, and thus results in lower accuracy. HidSim, despite providing consistent results doesn’t fully leverage the power of token-level analysis. Our intuition behind this is because it relies on average token representations rather than token-by-token analysis, which is overcome by our method.

\subsection{Experiment on KV Retrieval}
\begin{table}[h!]
\centering
\begin{tabular}{lccc}
\toprule
\textbf{KVData} & Llama-3 & Mistral  \\
\midrule
\concite ($\beta$=8) & 82.80  & 80.80   \\
\concite ($\beta$=64) & 100  & 100   \\
\hidsim & 90.98 & 73.35 \\
\shapvalue & \textbf{100} &\textbf{100}  \\
\midrule 
\multicolumn{2}{l}{\embsim OpenAI text embedding} & 57.81 \\
\bottomrule
\end{tabular}
\caption{Attribution Accuracy on KV Retrieval dataset with 500 examples. $\beta$ is the number of random samples used for attribution in \concite.}
\label{tab:kv}
\end{table}


The results in Table \ref{tab:kv} show that ConCite with $\beta$=64 and TokenShapley both reached perfect accuracy for both Llama-3 and Mistral. HidSim performed moderately well. In contrast, the off-the-shelf text embedding method (EmbSim) only achieved 57.81\% accuracy. We exclude the Yi model from our evaluation due to its 4K context length limit, while the dataset requires approximately 5K tokens. This dataset is a simple task for non-LLM algorithms because a naive exact match can achieve 100\% accuracy. We add this experiment to check if an attribution algorithm can also get 100\% accuracy. But we can see that similarity based methods fail to find all the true values.

\subsection{Experiment on Natural Question}




Figure \ref{fig:nq} shows the performance of various attribution methods on the Natural Questions dataset. 1000 examples are selected from the dataset. Overall, TokenShapley performs similarly well with ConCite when $k<5$ and it performs consistently better when $k\geq 5$, which is likely due to its token-level attribution that better captures subtle contributions in complex documents. In contrast, methods like HidSim and EmbSim rely on overall hidden state or embedding similarities, and thus it missed fine-grained token contributions. Moreover, we  see that TokenShapley works similarly with \concite, yet it improves performance when more sentences are selected, indicating its advantage in capturing broader contextual information. Table \ref{tab:f1} reports the precision, recall, and F1 scores for various methods when $k=5$. \shapvalue outperforms the other baselines, indicating its superior effectiveness in identifying the correct attribution.


\begin{table}[h!]
\centering
\begin{tabular}{lcccc}
\toprule
\textbf{NQ} & Precision & Recall &F1 \\
\midrule
\concite-8  & 0.268   & 0.425 &     0.329 \\
\concite-64 & 0.334 &   0.514 &     0.405\\
\concite-256 &  0.339 &   0.523 &     0.412\\
\hidsim &  0.163 &   0.197 &     0.178  \\
\embsim & 0.084 &   0.107  &     0.094 \\
\shapvalue  & \textbf{0.344} &   \textbf{0.528} &    \textbf{0.417} \\
\bottomrule
\end{tabular}
\caption{Precision, recall and F1 on Natural Question dataset with 1000 examples. 8, 64, 256 are the number of random samples used for attribution in \concite. Here we select top 5 candidates, $k=5$.}
\label{tab:f1}
\end{table}

\subsection{Experiment on CNN Dailymail}
\begin{table}[h!]
\centering
\begin{tabular}{lccc}
\toprule
\textbf{CNN Dailymail} & Llama-3 & Mistral  \\
\midrule
\concite ($\beta$=8) &  0.88 &  0.91  \\
\concite ($\beta$=64) & \textbf{1.38}  & \textbf{1.48}   \\
\hidsim & 0.72 & 0.30 \\
\shapvalue & 1.01 & 1.33 \\
\midrule 
\multicolumn{2}{l}{\embsim OpenAI text embedding} & 0.46 \\
\bottomrule
\end{tabular}
\caption{Log Probability Drop on CNN Dailymail dataset with 1000 examples. $\beta$ is the number of random samples used for attribution in \concite.}
\label{tab:cnn}
\end{table}


Note that \concite directly optimizes the log probability drop metric, which explains its strong performance when $\beta = 64$. As shown in Table \ref{tab:cnn}, TokenShapley achieves comparable results to \concite and even outperforms \concite with $\beta = 8$. In contrast, HidSim results in much smaller drops (0.72 for Llama-3 and 0.30 for Mistral), and the off-the-shelf OpenAI text embedding similarity (EmbSim) performs poorly, with a drop of only 0.46. These results indicate that TokenShapley effectively identifies truly influential context, whereas simple similarity-based methods fail to capture attribution quality.

\section{Discussion on Computational Cost}
Our method involves two main computational components: (1) offline datastore construction, and (2) runtime KNN-Shapley inference.

\textbf{Datastore Construction.} This step is performed once and offline. For each token in the context dataset, we cache its prefix embedding, resulting in a datastore whose size is linear in the number of context tokens. For example, constructing the datastore for the NaturalQuestions dataset requires approximately 100KB of memory per example (computed as embedding dimension 768 × number of tokens). Importantly, we only store one Wikipedia page per question, rather than the full Wikipedia corpus, which greatly reduces the memory footprint. \textbf{Runtime Inference.} At inference time, the KNN-Shapley computation retrieves top-$K$ neighbors from the datastore to assess token-level attribution. When $K=1$, the computational complexity is $\mathcal{O}(n)$, where $n$ is the number of tokens in the input context. In general, for $K > 1$, the complexity becomes $\mathcal{O}(n^2)$, though $n$ remains small in practice (e.g., <512 tokens). We further reduce overhead by precomputing and storing datastore embeddings.

Given the small per-example memory cost and the offline nature of the datastore construction, the overall computational and memory overhead of our method is minimal. In addition, the modest datastore size allows it to be stored efficiently in memory or on low-cost cloud storage, making our approach practical for real-world applications.

\section{Conclusion}
In this work, we introduced TokenShapley, a novel token-level attribution method that integrates Shapley value-based data attribution with KNN-based retrieval, leveraging a precomputed datastore for efficient contextual retrieval. By computing Shapley values to quantify token importance, TokenShapley enables a more precise and interpretable attribution of language model outputs. 


\section{Limitations}


One limitation of TokenShapley is the additional computational overhead required to create the datastore. However, this process only needs to be performed once and can be done offline. Future research may explore ways to enhance the efficiency of datastore creation. Another limitation is that the current method does not extend to text-to-image generation. TokenShapley relies on exact token-to-token matching, but identifying such correspondences in images remains an open challenge. We leave this for future work. One potential risk of TokenShapley is that datastore creation may expose user information, potentially compromising privacy. Future work should focus on developing privacy-preserving techniques for context attribution.


\bibliography{ref}

\appendix

\label{sec:appendix}



\section{Details for Weighted KNN Shapley}
\label{app:weight-knn}

\begin{table}[h]
\begin{center}
\caption{Table of Notation}\label{tab:notation}
\resizebox{1 \linewidth}{!}
{
\begin{tabular}{|cl|}\hline
$D$ :& Dataset, i.e. the context.\\
$S$ : & A subset of $D$. \\
$\mathcal{C}$ : & Datastore cache. \\
$z$ :& Data point in the dataset \\
& i.e. a token in the context. \\
$(x, y)$: &  (feature, label) pair for $z$.\\
${z} \to ({x}, {y})$: & The mapping from token to \\ &(feature, label) pair. \\
$\hat{z} , \hat{x}, \hat{y}$: & Token, feature, label in response. \\
\hline
$\U(\cdot)$: & The utility function for Shapley value.\\
$\phi_z(\U)$ : & The Shapley value for a data point $z$.\\
$h(\cdot)$ : & Hidden state function.\\
$s(\cdot, \cdot)$ :& Similarity function.\\
\hline
$N$ : & The number of points in $D$.\\
$K$: & Select the nearest $K$ neighbors. \\
$\ks$: &$\min(K, |S|)$\\
$\xsj$: & The j-th closest prefix to $\hat{x}$ in $S$\\
$M$ :& The number of data points used for \\
& updating Shapley in each KNN.\\
\hline
\end{tabular}
}
\end{center}
\end{table}

We define the weights in Equation \ref{eq:knnscore} in a more compact form,
\begin{align}
   \omega_j^{(S)}(\hat{x}):= \begin{cases}
       s(h(\hat{x}), h(x_j^{(S)})), & y_j^{(S)} = \hat{y} \\
       -s(h(\hat{x}), h(x_j^{(S)})), & y_j^{(S)} \neq \hat{y}
   \end{cases}
\end{align}
Then Equation \ref{eq:knnscore} becomes
\begin{align}
&\U(S; \hat{x}_t) = \mathbb{I} \left[ \sum_{j=1}^{\ks} \omega_j^{(S)} \geq 0 \right]
\end{align}

For a data point $z_i = (x_i, y_i)$, let 
\begin{align}
    \omega_i = s(h(\hat{x}), h(x_i))
\end{align}

Then the three conditions in Equation \ref{eq:diff} have the following expressions.
\begin{align}
    &\text{Cond}_{KNN} := z_i \text{ is within }\\
    \nonumber
   &  K \text{ nearest neighbors of }
    \xval \text{ among } S \cup \{z_i\} 
\end{align}

\begin{align}
    &\text{Cond}_{0\to1} := \\
    \nonumber
    &\begin{cases}
        \sum_{z_j \in S} \omega_j^{(S)}(\hat{x})  \in [-{\omega}_i, 0) , \quad \text{if } |S| \leq K - 1 \\
        \sum_{j=1}^{K-1} \omega_j^{(S)}(\hat{x}) \in \left[-\omega_i, -\omega_K^{(S)}(\hat{x}) \right) , \text{if } |S| \geq K
    \end{cases}
\end{align}

\begin{align}
    &\text{Cond}_{1\to0} :=\\
    \nonumber
    &\begin{cases}
        \sum_{z_j \in S} \omega_j^{(S)}(\hat{x})  \in [0, {\omega}_i),\quad \text{if } |S| \leq K - 1 \\
        \sum_{j=1}^{K-1} \omega_j^{(S)}(\hat{x})  \in \left[-\omega_K^{(S)}(\hat{x}) , \omega_i\right), \text{if } |S| \geq K
    \end{cases}
\end{align}


\section{Implementation Details}
\label{app:setting}

\begin{table}[ht]
\centering
\caption{Experimental Results for Different Parameter Settings on Verifiability Dataset}
\label{tab:results_hidsim}
\fontsize{10pt}{12pt}\selectfont  
\setlength{\tabcolsep}{4pt} 
\begin{tabular}{@{}lcccc@{}}  
\toprule
\textbf{Model} & \multicolumn{2}{c}{$K = 15$} & \multicolumn{2}{c}{$K = 25$} \\ 
                       & $W = 10$ & $W = 15$ & $W = 10$ & $W = 15$ \\ 
\midrule
\textbf{Llama-3}      & 69.2 & 70.8 & 69.7 & 70.8 \\
\textbf{Mistral-7B}    & 69.2 & 69.7 & 71.3 & 69.7 \\
\textbf{Yi-6B}             & 61.0 & 60.5 & 61.5 & 60.0 \\
\bottomrule
\end{tabular}
\end{table}

We implemented the baseline method HidSim \citep{phukan2024peering}  for text attribution following their paper. Table \ref{tab:results_hidsim} presents the accuracy results for different parameters demonstrating this baseline's influence on model performance on Verifiability dataset. The method involved extracting hidden layer representation for attributing text spans. First, each relevant text passage, with questions and summaries, was combined, tokenized, and passed through the model to provide intermediate representations. We chose embeddings from the middle layer, as it provides a balanced representation of the syntactic and semantic information from the model.  Second, for each span extracted from the summaries, we identify the annotated text passages and extract its indexes and texts. The indexes are used as ground truth, while texts were used to generate an embedding of the span using the same model. Next, we computed cosine similarity scores between these span embeddings and embeddings of the individual tokens within the passages. Then we identify the top-$K$ tokens with the highest similarity score as anchor tokens, following the paper. Finally, for each anchor token, we created a window of  tokens centering (having a size of $W$) on these anchor tokens. For each window, we compute the similarity score with respect to the span embedding and choose the window with maximum similarity. We use its index as the predicted location of the attributed text and compare it with the ground truth index extracted earlier.

It must be mentioned that the parameters top-$k$ and window size are chosen by design choice, as the paper \citep{phukan2024peering} doesn't mention these parameters configurations.  Given the middle layer provides a balanced representation of the texts, we chose a higher value of top-$K$  to ensure sufficient anchor tokens are chosen.  Moreover, we chose a value of moderate window size that ensures attributing the span correctly for all examples, including those having broader contexts. We tuned the parameters top-$K$  and window size and investigated its performance on model accuracy. For example, increasing the window size from 10 to 15 tokens improved the Meta-llama model's accuracy by 2\%, while increasing the top-$K$  from 15 to 25 enhanced the performance by 1.5\%. These results indicate that a broader selection of anchors can facilitate accurate attribution of the spans.

\section{Case Study}

\begin{tcolorbox}[colback=blue!10, colframe=blue!80!black, title=\textbf{Context}]

The following is a list of the top National Football League (NFL) quarterbacks in wins. In the NFL, the quarterback is the only position that is credited with records of wins and losses.
\newline
\newline
\textcolor{blue}{\hl{Active quarterback Tom Brady holds the records for most wins with 220, most regular season wins with 195, and most postseason wins with 25, as of Week 16 of the 2017 NFL season}}. 
\newline
\newline
\hl{Having played the entirety of his career with the New England Patriots, each of Brady 's win records also apply to wins with a single team.}
\newline
\newline
Among retired players, the record for most wins is held by Peyton Manning with 200. \textcolor{blue}{In his final professional game, Manning set the then - record for wins, surpassing Brett Favre who retired with 199 wins.} Other previous record - holders include John Elway (162), Fran Tarkenton (130), and Johnny Unitas (124). Otto Graham holds the record for the highest winning percentage with a minimum of 35 wins at. 788 (61 wins to 16 losses).

\end{tcolorbox}


\begin{tcolorbox}[colback=blue!10, colframe=blue!80!black, title=\textbf{Query}]

Who had the most wins in the NFL?

\end{tcolorbox}


\begin{tcolorbox}[colback=blue!5, colframe=blue!80!black, title=\textbf{Response}]

According to the given information, Tom Brady holds the record for the most wins in the NFL with 220 wins.

\end{tcolorbox}

In this section, we present an example of attribution results produced by \shapvalue on the Natural Questions dataset. The context provides background information about the NFL player Tom Brady. The question asks, "Who had the most wins in the NFL?" and the response correctly states that Tom Brady had the most wins. We apply \shapvalue to identify the most influential sentences (evidence) in the context for generating the response. The top two sentences selected by \shapvalue are shown in blue: the first receives a Shapley score of 16.8, and the second receives a score of 2.83. The yellow highlighted text corresponds to the ground-truth evidence labels provided by the dataset. We observe that both the precision and recall are 0.5. Notably, the sentence "... Tom Brady holds the records for most wins with 220, ..." is correctly identified by \shapvalue, demonstrating its effectiveness. However, the sentence "... record for wins ..." also receives a non-negligible score. We hypothesize that this is due to the high surface-level similarity between "record for wins" and "most wins", which may lead to confusion. Future work may improve attribution precision by incorporating additional filtering mechanisms to reduce such spurious matches.




\end{document}